\documentclass[letterpaper]{article} % DO NOT CHANGE THIS
\usepackage{aaai2026} 
\usepackage{times}  % DO NOT CHANGE THIS
\usepackage{helvet}  % DO NOT CHANGE THIS
\usepackage{courier}  % DO NOT CHANGE THIS
\usepackage[hyphens]{url}  % DO NOT CHANGE THIS
\usepackage{graphicx} % DO NOT CHANGE THIS
\usepackage{natbib}  % DO NOT CHANGE THIS AND DO NOT ADD ANY OPTIONS TO IT
\usepackage{caption} % DO NOT CHANGE THIS AND DO NOT ADD ANY OPTIONS TO IT
\usepackage{algorithm}
\usepackage{algorithmic}
 \usepackage{amsfonts} 
 \usepackage{amsmath}
 \usepackage{multirow}
\usepackage[table,xcdraw]{xcolor}
\usepackage{array}
\usepackage{booktabs}
\usepackage{makecell}
\usepackage{newfloat}
\usepackage{listings}
\DeclareCaptionStyle{ruled}{labelfont=normalfont,labelsep=colon,strut=off} % DO NOT CHANGE THIS
\floatstyle{ruled}
\newfloat{listing}{tb}{lst}{}
\floatname{listing}{Listing}
\title{S\textsuperscript{3}LoRA: Safe Spectral Sharpness–Guided Pruning in Adaptation of Agent Planner}
\author{
    Shuang Ao,
    Gopal Rumchurn
}
\affiliations{
    \textsuperscript{\rm 1}School of Electronics and Computer Science \\
    University of Southampton\\
    Southampton, UK\\
    s.ao@soton.ac.uk, sdr1@soton.ac.uk
}

\usepackage{bibentry}
\begin{document}

\nocopyright
\maketitle

\begin{abstract}
 % Adapting Large Language Models (LLMs) using parameter-efficient techniques like LoRA has become a common practice for enabling complex capabilities in LLM-based agents. Regardless of its effectiveness, such adaptation can unintentionally degrade safety alignment, leading to unstable or unsafe behavior. We propose S\textsuperscript{3}LoRA, a lightweight, data-free framework that enhances safety by inspecting and selectively pruning the fine-tuned LoRA weight updates. Our method introduces Magnitude-Aware Spherically Normalized SVD (MAS-SVD) to robustly analyze the spectral properties of LoRA layers and defines a Spectral Sharpness Index (SSI) to identify and prune layers with concentrated, potentially risky updates. Extensive experiments and ablation studies demonstrate that our proposed method improves safety metrics while maintaining or enhancing task performance and reducing inference cost. These results position our method as a scalable and efficient solution for deploying safer and more reliable agent planners in real-world applications.

Adapting Large Language Models (LLMs) using parameter-efficient fine-tuning (PEFT) techniques such as LoRA has enabled powerful capabilities in LLM-based agents. However, these adaptations can unintentionally compromise safety alignment, leading to unsafe or unstable behaviors, particularly in agent planning tasks. Existing safety-aware adaptation methods often require access to both base and instruction-tuned model checkpoints, which are frequently unavailable in practice, limiting their applicability. We propose S\textsuperscript{3}LoRA (Safe Spectral Sharpness–Guided Pruning LoRA), a lightweight, data-free, and model-independent framework that mitigates safety risks in LoRA-adapted models by inspecting only the fine-tuned weight updates. We first introduce Magnitude-Aware Spherically Normalized SVD (MAS-SVD), which robustly analyzes the structural properties of LoRA updates while preserving global magnitude information. We then design the Spectral Sharpness Index (SSI), a sharpness-aware metric to detect layers with highly concentrated and potentially unsafe updates. These layers are pruned post-hoc to reduce risk without sacrificing task performance. Extensive experiments and ablation studies across agent planning and language generation tasks show that S\textsuperscript{3}LoRA consistently improves safety metrics while maintaining or improving utility metrics and significantly reducing inference cost. These results establish S\textsuperscript{3}LoRA as a practical and scalable solution for safely deploying LLM-based agents in real-world, resource-constrained, and safety-critical environments. The code is available at \url{https://github.com/AoShuang92/S3_LoRA}.

\end{abstract}


\section{Introduction}

% Agent
% Agent safety
% LLM based Agent planning, safety
% LLM safety/Pruning
% Safety solely based on LoRA/single model Weight: A model weight itself can reflect the xxxxxxxxx SVD/Spectral. However, not for agent planning, technical: SVD slower, outliers,  SP SVD

% In this paper,.........

% summary

\begin{figure}[!ht]
\centerline{\includegraphics[width=0.5\textwidth]{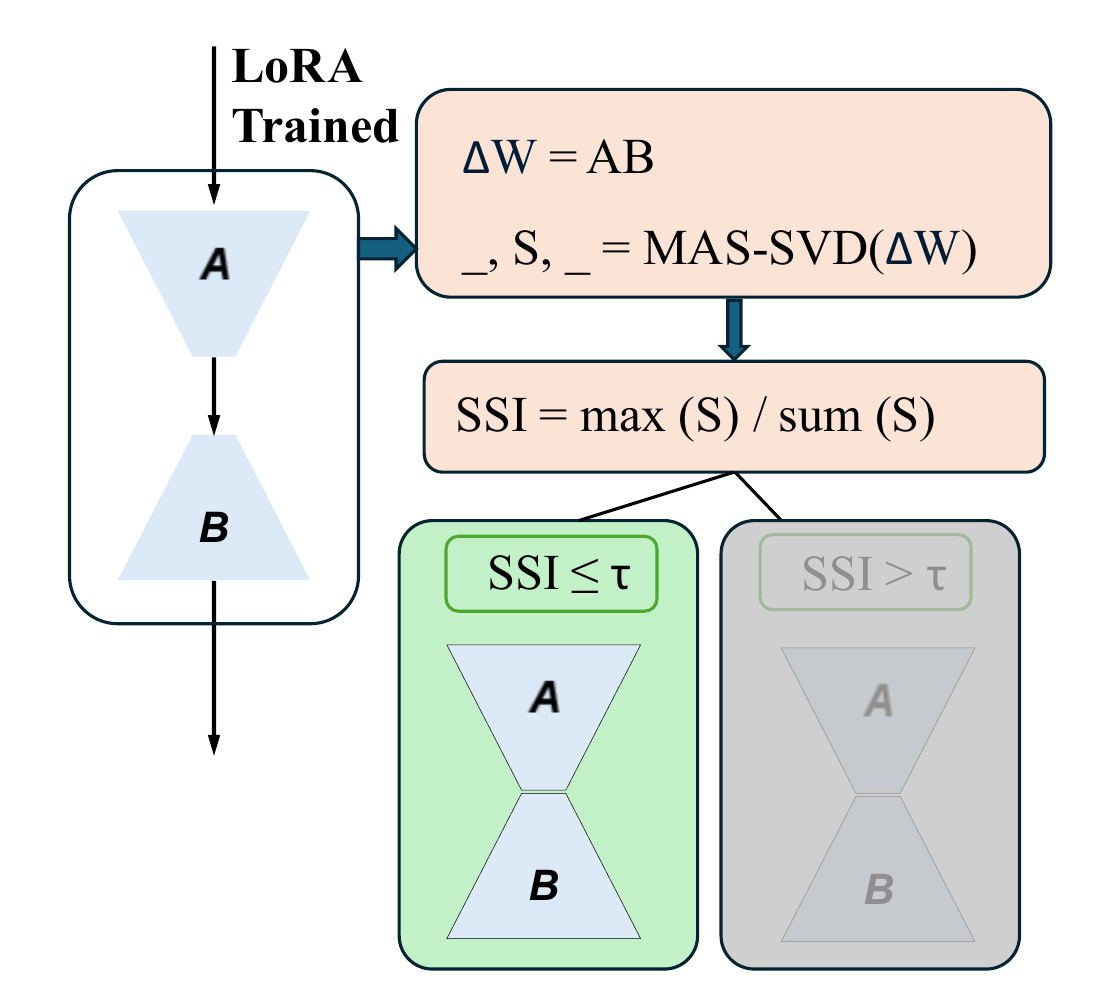}}
\caption{ Overview of S\textsuperscript{3}LoRA method. Each LoRA update $\Delta W$ is decomposed using MAS-SVD to obtain spectral values. The Spectral Sharpness Index (SSI) is then computed, and layers with high SSI scores are pruned to suppress unsafe updates while preserving model utility.}
\label{fig:overview}
\end{figure}

Large Language Models (LLMs) have demonstrated strong capabilities in reasoning, generalization, and instruction-following across diverse natural language tasks~\cite{touvron2023llama, wei2024assessing, achiam2023gpt, bubeck2023sparks}. Building on these strengths, LLM-based agents have been developed to perform more complex tasks by interacting with external tools, humans, and the physical world~\cite{wang2024survey, xi2025rise, xie2023openagents}. Planning, which involves formulating coherent and context-aware sequences of actions, remains a key challenge for LLM-based agents, as current approaches often rely on rigid or overly simplified assumptions. Poor planning can result in hazardous behavior, redundant or looping actions, and incomplete task execution, leading to safety concerns and significant computational inefficiencies~\cite{hu2025agentgen,zeng2023agenttuning,xu2023rewoo}.

Agent planning often requires fine-tuning pretrained LLMs to generate context-aware and goal-directed action sequences, and parameter-efficient fine-tuning (PEFT) methods such as Low-Rank Adaptation (LoRA)~\cite{hu2022lora} are commonly used due to their efficiency and effectiveness. These methods enable models to better follow task-specific instructions while minimizing computational overhead, which is essential for planning tasks that involve complex decision-making over extended action trajectories. However, recent studies~\cite{qi2023fine,yang2023shadow,zhan2023removing} have indicated that LoRA fine-tuning can inadvertently compromise the safety alignment properties inherent in pretrained LLMs, even when applied using benign datasets. Although LoRA effectively improves performance on specific downstream tasks, this improvement can coincide with a degradation of the safety property embedded in the original model. Weakening safety alignment can result in diminished generalization, increased risk of overfitting, and catastrophic forgetting. Recent methods~\cite{hsu2024safe, ao2025safe} for improving safety alignment via arithmetic interventions, rely on access to both base (e.g., LLaMA2-7B) and instruction-tuned (e.g., LLaMA2-7B-Chat) versions of models to identify parameter regions associated with unsafe behavior. However, such reliance of paired base and instruct version of LLMs poses a significant limitation, as many widely used LLMs only publicly release either base or instruct version. For instance, models in the GPT family such as GPT-2, GPT‑NeoX‑20B, and GPT‑J‑6B, as well as IBM’s Granite 4.0, have only released base checkpoints. Claude2 has not released any official model weights, with only community-generated fine-tuned variants like Claude2-Alpaca available. Similarly, domain-specific models such as LLaVA-Med~\cite{li2023llava} are built on top of LLaVA~\cite{liu2023visual}, which itself is fine-tuned from LLaMA models, making the original base-instruct incompatible. These limitations hinder the use of paired-model methods for safety intervention and pose challenges for building reliable, planning-capable agents, as undetected vulnerabilities in fine-tuned models can lead to unsafe behaviors during execution.

The reliance on base-instruct models becomes more critical in agent planning, where LLMs are often fine-tuned through multiple intermediate stages such as modality alignment, synthetic data generation, or trajectory-based tuning~\cite{chen2025atlas, song2024agentbank, hu2025agentgen}. Although this process enhances task performance, it can introduce cumulative shifts that weaken the safety alignment of the original model. Consequently, the safety guarantees of base LLMs may not carry over to their downstream agent variants, and the effects of this misalignment remain largely unexamined.

% Among these techniques, Singular Value Decomposition (SVD) is widely adopted, as it captures low-rank structures and reveals how parameter distributions evolve during training.
Studies have demonstrated that the trained weights of a model can reflect its internal behavior through spectral decomposition, without requiring access to external data or pretrained weights. 
% For example, prior work~\cite{bhandari2025forecasting, yunis2024approaching, barsbey2025interaction} has demonstrated that trained weight matrices, without the need for external data or pretrained checkpoints, can reveal underlying learning dynamics of neural networks through spectral analysis. These studies apply singular value decomposition to investigate phenomena such as transferability, catastrophic forgetting, and generalization. Notably, studies~\cite{bhandari2025forecasting, barsbey2025interaction} also introduce pruning strategies by truncating singular components associated with redundant or interfering directions. However, these approaches are primarily developed for discriminative settings such as image classification, and do not directly address the challenges of generative modeling. 
Recent work~\cite{wang2025svd, li2025adasvd} have extended spectral analysis to LLMs by identifying unsafe or misaligned directions in the model's parameter or representation space. These methods typically rely on auxiliary calibration datasets, multiple model comparisons, or access to hidden states, which increases computational overhead and limits their use in lightweight or constrained environments. In contrast, our goal is to develop an efficient, training-free diagnostic method that operates solely on the LoRA updated weights, without requiring access to a base model, its instruction-tuned counterpart, or any external data.

We propose Safe Spectral Sharpness-Guided Pruning LoRA (S\textsuperscript{3}LoRA), a post-hoc, data-free framework that identifies and removes potentially unsafe LoRA updates by analyzing only the fine-tuned weights. Central to our approach is Magnitude-Aware Spherically Normalized SVD (MAS-SVD), a spectral decomposition method that enhances robustness to outliers, reduces memory and computation, and preserves global magnitude information. Using MAS-SVD, we define the Spectral Sharpness Index (SSI) to measure the concentration of updates along dominant directions, where higher values indicate sharper and potentially unstable changes. Layers with the highest SSI scores are pruned to mitigate safety risks. An overview of this process is shown in Figure~\ref{fig:overview}. Our main contributions are as follows:

\begin{enumerate}

\item We propose Safe Spectral Sharpness-Guided Pruning LoRA (S\textsuperscript{3}LoRA), a pruning-based safety alignment strategy that removes potentially unsafe LoRA updates using a spectral sharpness criterion, without requiring additional data or retraining.

\item We introduce Magnitude-Aware Spherically Normalized SVD (MAS-SVD), a lightweight decomposition method that preserves global magnitude while being robust to outliers. Based on MAS-SVD, we define the Spectral Sharpness Index (SSI) to quantify concentrated and potentially unsafe parameter updates, which guides our pruning strategy.

\item By conducting extensive experiments and evaluations along with comprehensive ablation studies, we demonstrate that: 
    \begin{enumerate}
        \item S\textsuperscript{3}LoRA outperforms state-of-the-art (SOTA) safety alignment techniques, demonstrating the effectiveness of MAS-SVD in identifying risky layers;
        
        \item Safe Pruning approach significantly reduces computational overhead while preserving both performance and safety alignment;
        
        \item Our method strengthens model reliability by suppressing unsafe or inconsistent outputs.
    \end{enumerate}
    
\end{enumerate}

\section{Related Work}
% LLM Safety: SP-SVD
% LLM Pruning
% LLM based Agent Planner and Solver
% LLM based Agent Planning
% LLM based Agent Safety
% For example, SafeLoRA introduces a decomposed adapter structure that isolates unsafe directions in LoRA parameter space using base-instruct model pairs, enabling controlled updates to mitigate harmful generations. Similarly, SPLoRA learns a safety-preserving low-rank subspace by comparing base and instruct models, and constrains finetuning updates to remain within this safe subspace. 

% For example, ATLaS introduces a two-stage fine-tuning pipeline that first learns from full trajectories before refining critical decision steps; AgentBank aggregates large-scale multi-domain trajectories to pretrain agents before specialization; and AgentGen employs environment and task generation for curriculum-style intermediate training before planning fine-tuning. 

\subsection{Safety Agent Planner}
Recent advances in LLM-based agents have raised growing concerns about planning safety, as agents gain autonomy and interact with tools or the physical world. Agent Safety Alignment emphasizes the importance of defending against both unsafe user prompts and harmful tool outputs in multi-step agent planning~\cite{sha2025agent}. Safe-BeAl demonstrates that even task-successful plans by embodied agents can violate physical safety constraints, highlighting the need for safety-aware planning~\citep{huang2025framework}. AgentAlign reveals a growing tension between helpfulness and harmlessness as LLMs transition from passive assistants to agentic decision-makers~\citep{zhang2025agentalign}. These works demonstrate that safety in agent planning is a fundamental challenge. However, these approaches either rely on full model fine-tuning rather than PEFT methods like LoRA, introduce additional components or data for alignment, or incur significant computational overhead, making them less suitable for lightweight, modular integration into existing agent architectures.

\subsection{Spectral Decomposition}

Recent studies have leveraged spectral analysis to uncover critical insights into model weight dynamics. Singular values have been shown to encode task-relevant directions often overlooked during pruning, revealing spectral inconsistency across layers~\cite{staats2024small}; Yunis et.al~\citep{yunis2024approaching} explores temporal evolution of singular components in how models concentrate learning along dominant directions; and FARMS~\cite{hu2025eigenspectrum} utilize bias-corrected eigenspectrum estimation to improved the identification of heavy-tailed structures for better interpretability. However, these works primarily use spectral analysis for diagnostic or observational purposes, without offering actionable or structured interventions that translate these insights into model improvement. In contrast, methods grounded in statistical theory, such as Spherically Normalized SVD (SpSVD)~\cite{han2024robust}, improve robustness to outliers via row-wise normalization before decomposition. Barsbey et al.~\citep{barsbey2025interaction} show that compression techniques like neuron sparsity or spectral constraints can introduce sensitive directions, revealing a trade-off between compressibility and adversarial robustness. However, both approaches struggle to generalize to high-dimensional, task-specific representations typical of large-scale neural networks. In this work, we develop a spectral analysis method with both theoretical and empirical grounding, specifically designed to guide improvements in generative models.

% Spherically Normalized Singular Value Decomposition (SpSVD) is a recent and efficient algorithm designed for large-scale matrix analysis. It improves robustness to outliers by applying row-wise normalization before performing singular value decomposition. This normalization projects each row vector onto the unit hypersphere, preserving the algorithm’s robustness and insensitivity to outliers. However, SpSVD is proposed for theoretical analysis in large-scale matrix approximation and spectral estimation, with an emphasis on statistical guarantees in classical linear settings. By removing row-wise magnitude, it reduces the impact of high-norm outlier rows and allows singular value decomposition to focus on directional structure. In our work, we adapt SpSVD to better suit the characteristics of LLM fine-tuning, where both direction and magnitude of updates are informative.

\section{Methodology}

In this section, we propose Safe Spectral Sharpness–Guided Pruning LoRA (S³LoRA), a data- and training-free method for identifying and mitigating unsafe LoRA updates in agent planning. We first introduce Magnitude-Aware Spherically Normalized SVD (MAS-SVD), a robust spectral decomposition tailored to LoRA that preserves global magnitude while reducing memory and compute costs. From this, we derive the Spectral Sharpness Index (SSI) to measure the directional concentration of updates and prune LoRA layers that pose potential safety risks.

% Preliminaries:
% LLM Safety/alignment:
% SVD- spectral for model wieghts investigation
% equation: 
% S,h,v
% threshold
% LoRA: A(RHxW) 4096, B
% equ: cosine, DIEM, B*A\^T which obtain same dim as QHXW,K, V 4096x4096
% Redundant information, not singficant component

% Problem Statement/background:
% LLM based agent planning PEFT finetung of LLM. LLM can be making safe/harmful. Recent works attempted to fix/safety align it
% Comperssion the LORA matrix, principle component/significant component
% problems: unavailability of both base and instruct, cosine, dim, thresholding

% Proposed Method: Spectral based Safety Pruning
% %Safe Pruning LoRA Agent
% Obtain sigficant LoRA weights Dim
% SP-SVD spectral xx
% Thresholding
% Pruning

\subsection{Problem Statement}

Low-Rank Adaptation (LoRA) is a PEFT method that introduces trainable low-rank matrices into weight layers while keeping the original weights frozen, substantially reducing the number of trainable parameters for downstream tasks. For LLMs, the architecture comprises a stack of multi-head Transformer blocks, each containing attention sub-layers with distinct Q, K, V, and O linear projections. In our work, we use the term layer-wise to refer collectively to all Q, K, V, and O projection layers across all Transformer blocks.

For the $i$-th layer of a LLM, let the pretrained weight matrix be denoted by $W_0 \in \mathbb{R}^{d \times k}$. During LoRA fine-tuning, $W_0$ remains frozen, and the weight update is given by $W = W_0 + \Delta W = W_0 + AB$, where $\Delta W$ is the LoRA update. Given LoRA rank $r$, the matrices $A \in \mathbb{R}^{d \times r}$ and $B \in \mathbb{R}^{r \times k}$ are trainable low-rank adapters. 

Recent approaches such as SafeLoRA and SPLoRA leverage both a pretrained instruction-tuned model $W_0$ (e.g., LLaMA2-7B‑Chat), and its corresponding base model (e.g., LLaMA2-7B), denoted as $W_{\text{base}}$, to construct a safety-aligned subspace. LoRA updates are then projected into this subspace to identify and suppress unsafe directions. These methods require access to three sets of model weights: $W_0$, $W_{\text{base}}$, and the LoRA fine-tuned model $W$. Consequently, if any of these checkpoints are unavailable, the safety-aligned subspace cannot be reliably constructed, rendering the method inapplicable. 

Furthermore, if the pretrained model has undergone domain-specific fine-tuning, the alignment of the safety subspace between the original model and the final model parameters cannot be guaranteed. For instance, Med-LLaVA is fine-tuned based on LLaVA, which itself is derived from LLaMA. In this hierarchical fine-tuning scenario, the subspace constructed from LLaMA2-Chat and LLaMA2 does not capture the cumulative adaptations introduced through intermediate stages. A similar challenge arises with agent planning LLMs, which are often fine-tuned on domain-specific tasks or multimodal datasets. These modifications further deviate the model from its original alignment trajectory, making subspace-based methods insufficient to capture or enforce safety alignment properties in such specialized or cross-modal contexts. In this work, we focus exclusively on analyzing the LoRA update $\Delta W$ and treats it as a proxy for detecting potentially risky or anomalous layers that can affect model safety or alignment.

% Note that in some LLMs such as LLaMA, the dimensions $d$ and $k$ are same, but this has no impact on the applicability of our method. 

% \subsubsection{LLMs Safety Alignment}
\subsection{Safe Spectral Sharpness–Guided Pruning LoRA (S³LoRA)}

In this section, we introduce S\textsuperscript{3}LoRA, a post-hoc method for improving the reliability and efficiency of LoRA-adapted models by identifying and pruning potentially risky or redundant updates through robust spectral analysis. We first propose Magnitude-Aware Spherically Normalized Singular Value Decomposition (MAS-SVD), a fast and robust low-rank approximation technique that integrates directional robustness from spherical normalization with preserved magnitude information. This design yields stable and informative representations, making it particularly suitable for LLMs and LLM-powered agent planners. Building on MAS-SVD, we introduce the Spectral Sharpness Index (SSI), which is a metric that quantifies the sharpness of deviation in LoRA-updated weights across model layers. A higher SSI value reflects sharper deviations, as generalization error increases with sharpness or high spectral norms~\cite{yoshida2017spectral}. SSI functions both as a diagnostic tool for identifying layers with potentially unstable behavior and as a criterion for structured model pruning. Guided by this index, we selectively prune LoRA layers that are either safety-critical or contribute marginally to downstream performance.

\subsubsection{Magnitude-Aware Spherically Normalized Singular Value Decomposition (MAS-SVD)}

MAS-SVD first normalizes the weight matrix to ensure directional robustness, then extracts a stable low-rank structure resistant to outliers, and finally reintroduces magnitude information to recover meaningful scaling. This method enables accurate approximation of singular vectors and values while maintaining robustness in the complex and high-dimensional weight representations of LLMs.

The $i_{th}$ row of LoRA update matrix $\Delta W \in \mathbb{R}^{d \times k}$ is denoted by $\tilde{W}_{i,:} \in \mathbb{R}^{k}$. The row-wise normalization is written as: $\tilde{W}_{i,:}=\frac{\Delta W_{i,:}}{\left\|\Delta W_{i,:}\right\|_2+\varepsilon}, \quad
    \text{for } i = 1, \dots, d$, where $\left\|\Delta W_{i,:}\right\|_2$ is the Euclidean (l2) norm of the $i_th$ row vector; $\varepsilon$ is a small constant added for numerical stability to prevent division by zero. Sequentially, the $j_{th}$ column of the row-normalized matrix $\tilde{W}$ is $\tilde{W}_{:,j} \in \mathbb{R}^d$, and its column-wise normalization yields $
\hat{W}_{:,j} = \frac{\tilde{W}_{:,j}}{\left\| \tilde{W}_{:,j} \right\|_2 + \varepsilon}$.

The truncated singular value decomposition (SVD) is performed separately on the row-normalized matrix $\tilde{W}$ and the fully normalized matrix $\hat{W}$. Decomposing $\tilde{W}$ gives $\tilde{W} \approx \tilde{U} \tilde{S} \tilde{V}^\top$, and SVD on $\hat{W}$ yields $\hat{W} \approx \hat{U} \hat{S} \hat{V}^\top$. Throughout this paper, we use $U$, $S$ and $V$ to denote the left singular vectors, singular values (diagonal matrix), and right singular vectors respectively in any SVD, regardless of subscripts or the specific normalization.

To identify a robust low-rank structure of the matrix $\Delta W$, we define the candidate sets $\hat{U}^M$ and $\tilde{V}^M$ as the top-$M$ left and right singular vectors, obtained from the SVD of the fully normalized matrix $\hat{W}$ and the row-normalized matrix $\tilde{W}$ respectively. These candidate vectors span a set of rank-1 components used in the subsequent low-rank approximation of $\Delta W$. $M$ denotes the number of rank-1 components used to approximate $\Delta W$, which sets the target rank for the final low-rank reconstruction.

% At each step $m$, we search over all pairs $(u, v) \in U^M \times V^M$ and solve the following robust fitting objective: $\Delta W_m^{\mathrm{Sp}} = \arg\min_{u \in \hat{U}^M,\, v \in \tilde{V}^M,\, d \in \mathbb{R}} \left\| \Delta W - d u v^\top \right\|_1$.

At each step $m$, all pairs $(u, v) \in \hat{U}^M \times \tilde{V}^M$ are evaluated to solve the following robust fitting objective: $\Delta W_m^{\mathrm{Sp}} = \arg\min_{u \in \hat{U}^M,\, v \in \tilde{V}^M,\, d \in \mathbb{R}} \left\| \Delta W - d u v^\top \right\|_1$, where $\hat{U}^M$ and $\tilde{V}^M$ are the top-$M$ left and right singular vectors obtained from the SVD of the fully normalized matrix $\hat{W}$ and the row-normalized matrix $\tilde{W}$, respectively. This procedure is repeated iteratively with deflation, where previously selected components are subtracted from $\Delta W$, until $M$ components are extracted. The final approximation is then expressed as: $\Delta W_{final} \approx \sum_{m=1}^M \Delta W_m^{\mathrm{Sp}}$. We then perform singular value decomposition (SVD) on the final robust matrix $\Delta W_{final}$, yielding $\Delta W_{final} = U S V^\top$.

For LoRA update $\Delta W$, the magnitude of parameter changes encodes how strongly each layer contributes to model adaptation and potential safety misalignment. However, the spherical normalization process removes absolute scale information. To restore meaningful magnitudes after robust spectral decomposition, we propose to rescale the estimated singular values using the average row and column norms of the original (unnormalized) matrix $\Delta W$. Let the average row norm be denoted by $\bar{r}$:

\begin{equation}
\label{eq:row_norm}
    \bar{r} = \frac{1}{d} \sum_{i=1}^d \left\|\Delta W_{i,:} \right\|_2
\end{equation}

and the average column norm by $\bar{c}$:

\begin{equation}
\label{eq:col_norm}
    \bar{c} = \frac{1}{k} \sum_{j=1}^k \left\| \Delta W_{:,j} \right\|_2
\end{equation}

The magnitude-aware singular value matrix is then given by:

\begin{equation}
\label{eq:rescaled_S}
    S^{\prime} = S \cdot \bar{r} \cdot \bar{c}
\end{equation}

This scaling reintroduces the global magnitude information suppressed during normalization, preserving the semantic and functional significance of update strength across layers.

\subsubsection{Spectral Sharpness Index (SSI)}

To quantify the sharpness of weight deviation in each LoRA-updated layer, we propose the Spectral Sharpness Index (SSI), a scalar score derived from the rescaled singular values $S^{\prime}$ obtained in MAS-SVD. Intuitively, the largest singular value captures the dominant direction of change in the LoRA weight update. When it constitutes a large proportion of the total spectral energy (i.e., the sum of all singular values), it suggests a sharp, low-rank, and anisotropic perturbation. According to Wedin’s Theorem~\cite{wedin1972perturbation, o2023matrices}, such concentrated spectral shifts can lead to unstable deviations in the model’s output, underscoring their potential risk to safety and generalization. This concentration can correlate with instability or safety risks in LLM adaptation. Accordingly, SSI is defined as:

\begin{equation}
\label{eq:ssi}
\text {SSI} = \frac{\sigma_1^{\prime}}{\sum_{j=1}^h \sigma_j^{\prime}+\varepsilon}
\end{equation}

We retain the top‑$h$ singular values from the SVD for computing the SSI, where $\sigma_1^{\prime}$ denotes the largest singular value, $\sum_{j=1}^h \sigma_j^{\prime}$ is the total spectral magnitude, and $\varepsilon = 10^{-6}$ is a small constant added for numerical stability.

\subsubsection{SSI Guided LoRA Pruning}

After obtaining the Spectral Sharpness Index (SSI) for each LoRA-updated layer, we rank all layers in descending order according to their SSI values. We then prune the top-$\tau$ layers with the highest scores, as these are assumed to exhibit the most sharply concentrated updates. The intuition is that high spectral sharpness can signal directional overfitting or instability, reflecting inconsistent or overly aggressive updates during adaptation. By removing these layers, we aim to reduce such inconsistencies while preserving the remaining layers with lower SSI values, which tend to represent more balanced and generalizable adaptations. The remaining layers with lower SSI values are preserved, as they are more likely to reflect stable and generalizable updates. 

Specifically, we zero out the corresponding LoRA update $\Delta W = A B$, effectively nullifying the contribution of the LoRA path while retaining the frozen pretrained weight $W_0$. This selective pruning serves as a safety-aligned regularization strategy, mitigating the risk of sharp deviations while retaining the core adaptation capacity of the LoRA model.

\begin{equation}
\label{eq:prune}
\mathcal{R}(\Delta W)= 
\begin{cases}
\text{keep } \Delta W,  & \text{otherwise} \\
\text{prune } \Delta W, &  \text{if } \text {SSI} \in \text{top-}\tau
\end{cases}
\end{equation}

% \begin{equation}
% \label{eq:prune}
% \mathcal{R}(W)= 
% \begin{cases}
% \text{keep } W,  & \text{if } \text{SSI}^i \notin \text{top-}\tau \\
% \text{prune }, &  \text{if } \text{SSI}^i \in \text{top-}\tau
% \end{cases}
% \end{equation}

Since the pretrained weights $W_0$ remain frozen, the LoRA update $\Delta W$ serves as the sole source of adaptation. Thus, pruning based on excessively high SSI values directly removes unstable updates, enhancing overall robustness without degrading the pretrained model’s foundation.
% -----------------------------------------
\begin{table*}[!ht]
\centering
\caption{Results of different LoRA safety techniques on the Planner Instruction Tuning 2K dataset. The Planner setting evaluates performance solely based on planning quality, while the Solver setting assesses the full agent system, with results measured based on final task outcomes. HS (Harmfulness Score) and ASR (Attack Success Rate) are used to evaluate safety, whereas SR (Success Rate) and F1 score reflect the effectiveness of the agent's final output. Higher values ($\uparrow$) indicate better task performance, and lower values ($\downarrow$) indicate better safety. For clarity, all results except HS are reported as percentages.}
\label{tab:planner}
\scalebox{0.92}{
\begin{tabular}{c|c|cccc|cc|cc|cc}
\hline
& \multicolumn{1}{l|}{} 
& \multicolumn{6}{c|}{Planner: Instruction Tuning 2K Dataset} 
& \multicolumn{4}{c}{Solver} \\
\cline{2-12}
& \multicolumn{1}{l|}{} 
& \multicolumn{4}{c|}{Utility Metrics ($\uparrow$)} 
& \multicolumn{2}{c|}{Safety Metrics ($\downarrow$)} 
& \multicolumn{2}{c|}{HotpotQA} 
& \multicolumn{2}{c}{TriviaQA} \\
\cline{2-12}
\multirow{-3}{*}{Category} & \multicolumn{1}{l|}{} 
& BLEU & ROUGE & METEOR & AUARC 
& ASR & HS 
& SR & F1 
& SR & F1 \\
\hline

Zero-shot LLM & Baseline 
& 16.03 & 20.92 & 23.89 & 57.03 
& 3.65 & 1.95 
& 22.64 & 20.42 
& 52.33 & 41.82 \\
\hline

\multirow{2}{*}{\shortstack{Zero-shot\\ Agent}} 
& \begin{tabular}[c]{@{}c@{}}AgentLM\end{tabular} 
& 27.63 & 36.42 & 32.21 & 68.82 
& 2.93 & 1.89 
& 35.45 & 32.35 
& 64.46 & 53.65 \\
\cline{2-12} % ← Draws line from 2nd to 12th column only (preserves multirow)
& \begin{tabular}[c]{@{}c@{}}Agent-FLAN\end{tabular} 
&  28.36 & 37.48 & 33.65 &  70.52
& 2.85 &  1.76
&  39.62 &  35.46
& 68.64 & 60.17 \\
\hline

PEFT & LoRA 
& \textbf{56.87} & \textbf{69.89} & 70.76 & 89.70 
& 2.36 & 2.01 
& 43.36 & \textbf{41.28} 
& 72.54 & \textbf{65.21} \\
\hline

& SafeLoRA 
& 55.35 & 68.81 & 69.85 & 90.72 
& 1.62 & 1.42 
& 42.31 & 40.56 
& \textbf{73.16} & 65.04 \\
\cline{2-12}

& SPLoRA 
& 55.44 & 69.27 & 69.86 & 91.56 
& 1.57 & 1.31 
& 42.96 & 40.68 
& 72.89 & 64.92 \\
\cline{2-12}

\multirow{-3}{*}{\shortstack{PEFT \\ with Safety \\Alignment}} & \textbf{S\textsuperscript{3}LoRA } 
& 56.15 & 69.81 & \textbf{70.94} & \textbf{93.08} 
& \textbf{1.23} & \textbf{1.15} 
& \textbf{44.52} & 40.86 
& 72.96 & 65.02 \\
\hline
\end{tabular}}
\end{table*}

% -----------------------------------------

% -----------------------------------------

\begin{table}[!ht]
\centering
\caption{Results of LLaMA2-7B-chat with various LoRA techniques on the AgentInstruct dataset. All results except HS are reported as percentages.}
\label{tab:agentistruct}
\scalebox{0.92}{
\begin{tabular}{c|cc|cc}
\hline
\multicolumn{5}{c}{\textbf{AgentInstruct Dataset}} \\ \hline
                      & \multicolumn{2}{c|}{Utility Metrics ($\uparrow$)} & \multicolumn{2}{c}{Safety Metrics ($\downarrow$)} \\ \hline
                      & METEOR         & AUARC          & ASR            & HS              \\ \hline
Baseline                    & 12.13          & 57.25          &     22.14       &  2.38               \\ \hline
LoRA                  & \textbf{25.16 }         & 75.21          & 21.15           &   2.04              \\ \hline
SafeLoRA              & 24.96          & 79.82          & 17.39           &   1.95              \\ \hline
SPLoRA                & 25.12          & 81.57          & \textbf{15.74  }         &  1.76               \\ \hline
\textbf{S\textsuperscript{3}LoRA}                  & 25.04 & \textbf{83.08} & 16.34  &       \textbf{1.52 }         \\ \hline
\end{tabular}}
\end{table}

% -----------------------------------------
\begin{figure}[!ht]
\centerline{\includegraphics[width=0.52\textwidth]{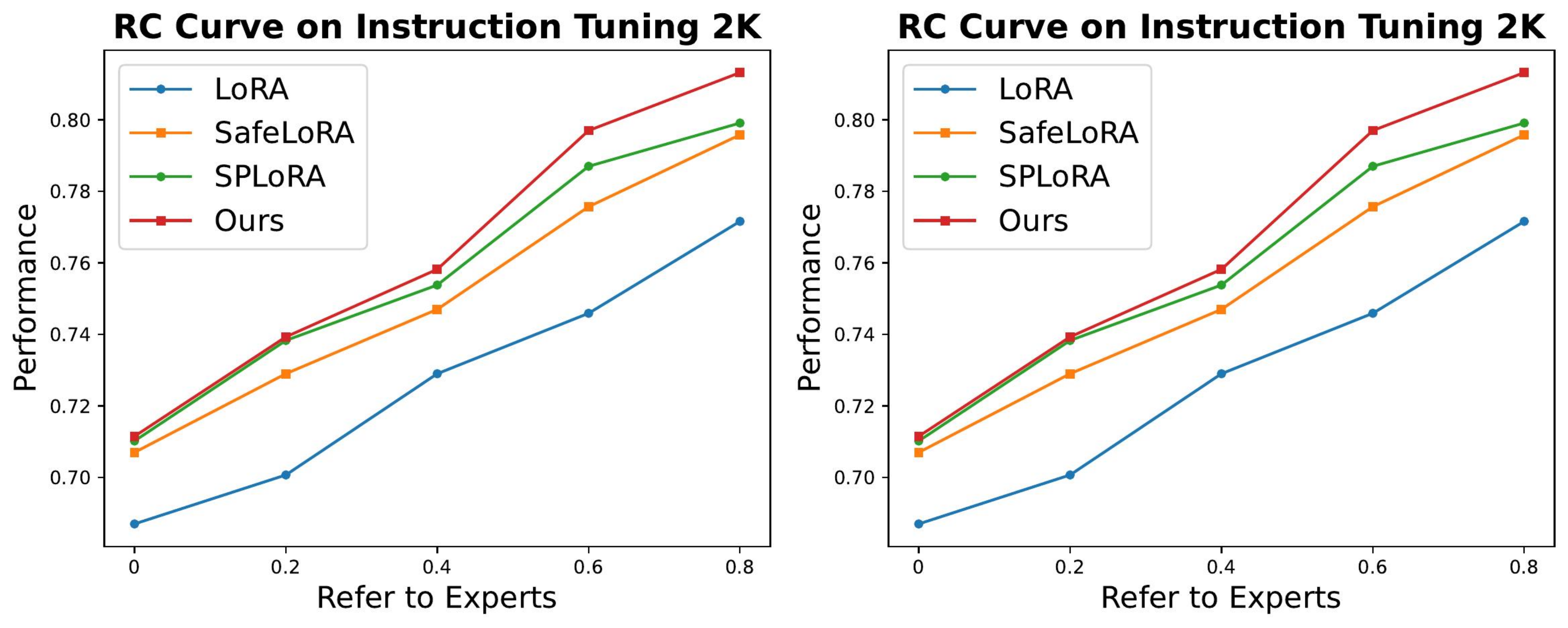}}
\caption{ The Risk-Coverage Curve compares LoRA, SafeLoRA, SPLoRA and our proposed S\textsuperscript{3}LoRA, with performance measured using the ROUGE-1 F1 score. The x-axis ("Refer to experts") represents the percentage of samples with the highest uncertainty scores. The left plot shows results for fine-tuning on Instruction Tuning 2K dataset with LLaMA2 model, and the right plot shows results for fine-tuning on Dialogue Summary dataset using the Gemma model.}
\label{fig:rcc}
\end{figure}

% -----------------------------------------

% -----------------------------------------
\begin{table*}[!ht]
\caption{Performance comparison of our methods against LoRA, SafeLoRA, Vaccine and SPLoRA on the Dialogue Summary and Alpaca dataset, using LLaMA2-7B-Chat and Gemma-7B-it models. HS (Harmfulness Score) and ASR (Attack Success Rate) are used to assess safety. Higher values ($\uparrow$) indicate better performance, and lower values ($\downarrow$) indicate better safety. For clarity, all results except HS are reported as percentages.}
\label{tab:llms}
\centering 
\begin{tabular}{c|c|c|ccc|cc} 
\toprule 
  \multirow{2}{*}{\makecell[c]{Dataset}} & \multirow{2}{*}{\makecell[c]{Model}} & \multirow{2}{*}{\makecell[c]{Method}} 
  & \multicolumn{3}{c|}{Utility Metrics ($\uparrow$)} & \multicolumn{2}{c}{Safety Metrics ($\downarrow$)} \\ 
  \cmidrule(lr){4-6} \cmidrule(lr){7-8}
  & & & ROUGE & METEOR & AUARC & ASR & HS \\ 
\midrule

\multirow{6}{*}{\makecell[c]{Dialogue\\Summary}} 
& \multirow{6}{*}{\makecell[c]{Gemma\\7B-it}} 
& LoRA        & 35.35 & 43.31 & 87.82 & 20.22 & 1.38 \\
& & Vaccine   & 36.24 & 43.21 & 85.32 & 7.53 & 1.17 \\
& & SafeLoRA  & 36.03 & 44.82 & 84.35 &8.20 & 1.23 \\ 
& & SPLoRA    & 36.91 & \textbf{44.96} & 87.32 & 6.07 & 1.12 \\ 
& & \textbf{S\textsuperscript{3}LoRA (Ours)} & \textbf{37.82} & 44.73 & \textbf{87.96} & \textbf{5.85} & \textbf{1.04} \\
\midrule

\multirow{6}{*}{\makecell[c]{Alpaca}} 
& \multirow{6}{*}{\makecell[c]{LLaMA2\\7B-Chat}} 
& LoRA       & 24.65 & \textbf{20.48} & 70.24 & 25.31 & 1.83 \\
& & Vaccine  & 24.45 & 19.86 & 67.54 & 11.23 & 1.34 \\
& & SafeLoRA & 24.22 & 20.45 & 66.82 & 7.54 & 1.15 \\  
& & SPLoRA   & 24.86 & 20.43 & 71.02 & 5.64 & 1.21 \\ 
& & \textbf{S\textsuperscript{3}LoRA (Ours)}  & \textbf{25.12} & 20.35 & \textbf{73.56} & \textbf{4.75} & \textbf{1.03} \\ 

\bottomrule
\end{tabular}
\end{table*}

% -----------------------------------------

% -----------------------------------------
\begin{table*}[!ht]
\caption{Impact of layer pruning threshold of SSI. Utility and safety metrics on the Instruction Tuning 2K dataset using the LLaMA2-7B-Chat model, evaluated under different pruning thresholds based on the number of pruned layers.}
\label{tab:layer}
\centering 
\begin{tabular}{c|c|c|ccc|cc} 
\toprule 
  \multirow{2}{*}{\makecell[c]{Model}} & \multirow{2}{*}{\makecell[c]{Pruned \\Layers}} & \multirow{2}{*}{\makecell[c]{Threshold\\Value}} 
  & \multicolumn{3}{c|}{Utility Metrics ($\uparrow$)} & \multicolumn{2}{c}{Safety Metrics ($\downarrow$)} \\ 
  \cmidrule(lr){4-6} \cmidrule(lr){7-8}
  & & & ROUGE & METEOR & AUARC & ASR & HS \\ 
\midrule

\multirow{4}{*}{\makecell[c]{LLaMA-2\\7B-Chat}} 
& 5 layers  & 0.44 & 69.06  & 68.54 & 91.27 & 1.35 &  1.32\\
& \textbf{10 layers} & 0.42 & 69.81 & 70.94 & 93.08 & 1.23 & 1.15 \\
& 15 layers   & 0.41 & 66.55 & 67.27 & 90.32  & 1.41 &  1.48\\ 
& 20 layers  & 0.39 & 65.74  & 66.34  & 89.25 & 1.54 &  1.67\\ 

\bottomrule
\end{tabular}
\end{table*}

% -----------------------------------------
\section{Experiments}

\subsection{Datasets and Baselines}

For the agent planning task, we use the Planner Instruction Tuning dataset~\cite{xu2023rewoo}, which combines task planning trajectories within the ReWOO (Reasoning WithOut Observation) framework. In addition, we use the AgentInstruct dataset~\cite{zeng2023agenttuning}, an instruction-tuning dataset containing approximately 1866 samples with high-quality interaction trajectories collected across six diverse real‑world tasks. Each dataset is split into 80\% for training and 20\% for testing. Model adaptation is performed using LoRA-based fine-tuning. 

We evaluate Planner Instruction Tuning dataset also as part of the full agent system with solver component, to assess overall execution performance for HotpotQA~\cite{yang2018hotpotqa} and TriviaQA~\cite{joshi2017triviaqa}.

To further validate our methodology, we also employ datasets for language generation tasks, specifically the Dialogue Summary~\cite{gliwa-etal-2019-samsum} and Alpaca~\cite{taori2023stanford} datasets. Evaluation is performed using 1,500 test samples for Dialogue Summary and 20\% of the total data for Alpaca.

For the agent planner, we use the LLaMA2-7B-Chat~\cite{touvron2023llama} model in both zero-shot and LoRA fine-tuning settings. In addition, we evaluate zero-shot performance using AgentLM (7B)~\cite{zeng2023agenttuning} and Agent-FLAN (7B)~\cite{chen2024agent}, both of which are fully fine-tuned variants of LLaMA2-7B. All models share the same architecture to ensure fair comparison. Our experiments also include the Gemma-7B-it~\cite{team2024gemma} and LLaMA2-7B-Chat models for general-purpose language modeling tasks.

We compare our proposed S\textsuperscript{3}LoRA with the following SOTA techniques:

\begin{enumerate}

\item LoRA~\cite{hu2022lora}: incorporates trainable low-rank matrices into pre-trained model weights to enable parameter-efficient fine-tuning (PEFT).

\item SafeLoRA~\cite{hsu2024safe}: enhances LoRA fine-tuning by projecting updates onto a safety-aligned subspace, aiming to suppress harmful outputs while retaining model utility.

\item Vaccine~\cite{huang2024vaccine}: proposes a perturbation-aware alignment strategy that strengthens robustness against harmful fine-tuning attacks. We evaluate this method in the context of language generation tasks.

\item SPLoRA~\cite{ao2025safe}: introduces a distance-guided pruning approach that detects and removes LoRA components detrimental to safety alignment, thereby reducing safety risks while maintaining task performance.

\end{enumerate}

\subsection{Evaluation Metrics}

In our experiments, we evaluate both utility and safety of the models using established metrics. Utility is assessed using BLEU, ROUGE-1 F1, and METEOR, which measure the similarity between model-generated responses and ground-truth references. We also include the Area Under the Accuracy-Rejection Curve (AUARC)~\cite{nadeem2009accuracy}, which measures the reliability of selective prediction. 

% To compute AUARC, each instance requires a binary correctness label and an associated uncertainty score. A response is considered correct if its ROUGE-L score with the reference exceeds 0.5~\cite{kuhn2023semantic, lin2023generating, ao2024css, kossen2024semantic}.  
% For uncertainty estimation, we use semantic entropy probes~\footnote{https://github.com/OATML/semantic-entropy-probes}, which compute uncertainty based on the sparsity of the output distribution~\cite{kossen2024semantic}.

Safety is evaluated using the Attack Success Rate (ASR) and Harmfulness Score (HS). An attack is considered successful if the model’s response lacks explicit refusal keywords, with the full list provided in the Appendix. Harmfulness is scored by GPT-4 on a 1–5 scale, where lower scores indicate safer outputs.

For evaluating agent performance with the solver component, We report the Success Rate (SR)~\cite{yehudai2025survey}, defined as the percentage of tasks the agent fully completes (i.e., achieving a reward of 1), and the token-level F1 score of the final output to assess generation accuracy at the level of individual tokens. To ensure a fair comparison across methods, all agents are paired with the same solver backend, GPT-3.5-Turbo, consistent with the settings used in the original benchmark.

\subsection{Implementation Details}

For our experiments, we use Hugging Face~\footnote{https://huggingface.co/} pre-trained LLaMA2-7B-Chat and Gemma-7b-it as baselines for zero-shot evaluation and LoRA fine-tuning. LoRA is applied to the "q\_proj," "k\_proj," "v\_proj," and "o\_proj" attention layers, using a fixed rank of 8 for all experiments. Fine-tuning is performed for 5 epochs with a batch size of 8. For all our experiments, we prune the top $\tau$ = 10 LoRA-updated layers with the highest SSI scores, as determined by our ablation study in Table~\ref{tab:layer}.

For the agent planning task, LLaMA2-7B-Chat is fine-tuned with a learning rate of 5e-5. For the Dialogue Summary task, Gemma-7B-it is fine-tuned with a learning rate of 5e-4. For the Alpaca dataset, LLaMA2-7B-Chat is again used with a learning rate of 5e-5. All experiments are conducted on two NVIDIA RTX A6000 GPUs, each with 48 GB of RAM.

\section{Results}

Table~\ref{tab:planner} summarizes performance on the Instruction Tuning 2K dataset, evaluating planning quality (Planner) and end-to-end execution (Solver). Zero-shot agents (AgentLM and Agent-FLAN) outperform the baseline but are outperformed by PEFT methods, with LoRA achieving the highest utility scores. Safety-aligned approaches (SafeLoRA, SPLoRA, and Ours S\textsuperscript{3}LoRA) slightly reduce utility but significantly improve safety, with our method achieving the lowest ASR and HS. In the Solver setting, our method obtains the highest success rate on HotpotQA and performs competitively on TriviaQA. Despite LoRA yielding the best F1, the risk-coverage curve in Figure~\ref{fig:rcc} (left) shows our method S\textsuperscript{3}LoRA provides more reliable behavior by effectively filtering unsafe or erroneous outputs.

Further evaluation on the AgentInstruct dataset using the LLaMA2-7B-chat model (Table~\ref{tab:agentistruct}) shows that our method S\textsuperscript{3}LoRA maintains strong utility while achieving the highest AUARC and lowest HS, demonstrating enhanced safety alignment.

To assess generalization to language generation tasks, we test on Dialogue Summary and Alpaca datasets using Gemma-7B-it and LLaMA2-7B-Chat, respectively. As shown in Table~\ref{tab:llms}, our method S\textsuperscript{3}LoRA delivers comparable utility to LoRA while consistently achieving the best safety scores across both datasets. The risk-coverage curve in Figure~\ref{fig:rcc} (right) further confirms improved robustness by prioritizing safer outputs under increasing risk thresholds.

\section{Ablation Studies}

We conduct a comprehensive ablation study alongside our main experiments to assess the effectiveness of S\textsuperscript{3}LoRA from multiple perspectives.

We evaluate the impact of layer pruning in S\textsuperscript{3}LoRA using the LLaMA2-7B-Chat model on the Instruction Tuning 2K dataset. Following the Spectral Sharpness Index (SSI), we rank all LoRA-updated layers by their SSI scores and prune the top $\tau$ layers with the highest values. As shown in Table~\ref{tab:layer}, pruning 10 layers achieves the best balance between utility and safety, yielding the highest AUARC and METEOR scores and the lowest ASR and HS. This configuration is used in all subsequent experiments, consistent with SafeLoRA~\cite{hsu2024safe} and SPLoRA~\cite{ao2025safe}, which also retain 10 projection layers.

% -----------------------------------------

\begin{table}[h]
    \centering
    \caption{Performance comparison of SVD, SpSVD and our MAS-SVD on Instruction Tuning 2K (IT2K) and Dialogue Summary (DS) datasets.}
    \label{tab:svd_comparison}
    \scalebox{0.95}{
    \begin{tabular}{lcccc}
        \toprule
        Dataset & Metric & SVD & SpSVD & \textbf{MAS-SVD} \\
        \midrule
        \multirow{3}{*}{IT2K} 
        & ROUGE ($\uparrow$) & 67.23 & 67.52 &  69.81  \\
        & METEOR ($\uparrow$) & 67.12 & 68.03 &  70.74 \\
        & ASR ($\downarrow$)  & 1.42  & 1.56  &  1.23 \\
        \midrule
        \multirow{3}{*}{DS} 
        & ROUGE ($\uparrow$)  & 24.02 & 23.21 & 25.12 \\
        & METEOR ($\uparrow$) & 18.35 & 19.28 & 20.32 \\
        & ASR ($\downarrow$)  & 6.03  & 5.46  & 4.75 \\
        \bottomrule
    \end{tabular}}
\end{table}
% -----------------------------------------

\begin{table}[ht]
\centering
\caption{Comparison of inference time and trainable parameters before and after pruning on the Instruction Tuning 2K dataset. "Per Sample" indicates the inference time per instance, and "\% Param" denotes the percentage of trainable parameters.}

\begin{tabular}{cccc}
\toprule
\textbf{Model} & \textbf{Method} & \textbf{Per Sample (s)} & \textbf{\% Param} \\
\midrule
\multirow{2}{*}{LLaMA2}  
& BS      & 1.56  & 100  \\  
& Pruned  & 1.21  & 1.12  \\
\midrule
\multirow{2}{*}{Gemma2}  
& BS      & 0.74  & 100  \\  
& Pruned  & 0.65  & 1.24  \\
\bottomrule
\end{tabular}

\label{tab:inference_time}
\end{table}

% -----------------------------------------
To evaluate the effectiveness of MAS-SVD, we replace it with SVD and SpSVD as the singular decomposition method in the S\textsuperscript{3}LoRA framework. As shown in Table~\ref{tab:svd_comparison}, MAS-SVD consistently outperforms both SVD and SpSVD across the Instruction Tuning 2K (IT2K) and Dialogue Summary (DS) datasets. These results highlight its effectiveness in maintaining a strong balance between task performance and safety across different domains.

We further evaluate the efficiency of our proposed method, by measuring per-sample inference time and the proportion of trainable parameters on the Instruction Tuning 2K dataset using the LLaMA2 model, and on the Dialogue Summary dataset using Gemma2. As shown in Table~\ref{tab:inference_time}, pruning reduces inference time by approximately 12–15\%, while dramatically decreasing the number of trainable parameters compared to the full baseline models. These results demonstrate that our approach not only improves safety and robustness but also offers clear computational benefits.
%-----------------------------------------------------------

\section{Conclusion}

In this work, we proposed S\textsuperscript{3}LoRA (Safe Spectral Sharpness–Guided Pruning LoRA), a lightweight, post-hoc method for improving the safety of LoRA-adapted language models, particularly in agent planning scenarios. Our approach leverages Magnitude-Aware Spherically Normalized SVD (MAS-SVD) to decompose LoRA updates and defines the Spectral Sharpness Index (SSI) to identify and prune layers with potentially unsafe sharp spectral deviations. This enables us to enhance robustness and reduce harmful behavior without access to base or instruction-tuned models, data, or retraining. Extensive experiments show that S\textsuperscript{3}LoRA improves safety alignment while maintaining strong task performance and lowering computational cost. While effective, the method involves a heuristic pruning threshold that may benefit from further tuning across different tasks, and it assumes a general correlation between spectral sharpness and risk, which might not fully capture domain-specific nuances. Future work includes exploring adaptive, performance-aware pruning strategies and integrating our method into broader alignment frameworks for safer LLM agents in complex, open-world environments.

% \appendix

% \section{Acknowledgments}

\bibliography{aaai2026}

\end{document}